\documentclass[10pt,twocolumn,letterpaper]{article}

\usepackage{cvpr}              % To produce the CAMERA-READY version
\usepackage{times}
\usepackage{epsfig}
\usepackage{graphicx}
\usepackage{amsmath}
\usepackage{amssymb}

\usepackage{booktabs}
\usepackage{csquotes}
\usepackage{tabularx}
\usepackage{subfiles}
\usepackage{float,xspace}
\newcolumntype{Y}{>{\centering\arraybackslash}X}
\newcolumntype{R}{>{\raggedright\arraybackslash}X}

\usepackage{comment}
\usepackage{algorithm,algorithmic,color}
\usepackage{boldline} 
\usepackage{dsfont}
\usepackage{color,soul}
\usepackage{balance}
\usepackage{multirow}
\usepackage{multicol}

\usepackage{diagbox}
\usepackage{xcolor}
\usepackage{enumitem}

\usepackage[normalem]{ulem}
\useunder{\uline}{\ul}{}

\newcommand{\cM}{\mathcal{M}\xspace}

\newcommand{\cR}{\mathcal{R}\xspace}
\newcommand{\cL}{\mathcal{L}\xspace}
\newcommand{\cC}{\mathcal{C}\xspace}

\newif\ifshowcomments
\newcommand{\done}[1]{}
\newcommand{\gunnardone}[1]{}
\newcommand{\shiyuandone}[1]{}
\newcommand{\vicentedone}[1]{}
\newcommand{\robinsondone}[1]{}

\showcommentstrue  % comment this out to hide comments
\ifshowcomments
    \newcommand{\gunnar}[1]{{\color{violet}[gunnar: #1]}}
    \newcommand{\shiyuan}[1]{{\color{orange}[shiyuan: #1]}}
    \newcommand{\vicente}[1]{{\color{purple}[vicente: #1]}}    
    \newcommand{\robinson}[1]{{\color{teal}[robinson: #1]}}
    
\else
    \newcommand{\gunnar}[1]{}
    \newcommand{\shiyuan}[1]{}
    \newcommand{\vicente}[1]{}    
    \newcommand{\robinson}[1]{}

\fi

\usepackage[pagebackref,breaklinks,colorlinks]{hyperref}

\def\cvprPaperID{*****} % *** Enter the CVPR Paper ID here
\def\confName{CVPR}
\def\confYear{2022}

\begin{document}

\title{Characterizing Video Question Answering with Sparsified Inputs}

\def\eg{\emph{e.g}\bmvaOneDot}
\def\Eg{\emph{E.g}\bmvaOneDot}
\def\etal{\emph{et al}\bmvaOneDot}

\author{Shiyuan Huang$^1$\ \
Robinson Piramuthu$^2$ \ \
Vicente Ordonez$^{2,3}$ \ \
Shih-Fu Chang$^1$\ \
Gunnar A. Sigurdsson$^2$ \\
$^1$ Columbia University\ \ 
$^2$ Amazon Alexa AI \ \ 
$^3$ Rice University
}

\maketitle

%%%%%%%%% ABSTRACT
\begin{abstract}
In Video Question Answering, videos are often processed as a full-length sequence of frames to ensure minimal loss of information. Recent works have demonstrated evidence that sparse video inputs are sufficient to maintain high performance. However, they usually discuss the case of single frame selection. In our work, we extend the setting to multiple number of inputs and other modalities. We characterize the task with different input sparsity and provide a tool for doing that. Specifically, we use a Gumbel-based learnable selection module to adaptively select the best inputs for the final task. In this way, we experiment over public VideoQA benchmarks and provide analysis on how sparsified inputs affect the performance. From our experiments, we have observed only $5.2\% {-} 5.8\%$ loss of performance with only $10\%$ of video lengths, which corresponds to $2{-}4$ frames selected from each video. Meanwhile, we also observed the complimentary behaviour between visual and textual inputs, even under highly sparsified settings, suggesting the potential of improving data efficiency for video-and-language tasks. 
\end{abstract}

%-------------------------------------------------------------------------

\section{Introduction}
\label{sec:intro}

\begin{figure}[h]
    \centering
    \includegraphics[width=\linewidth]{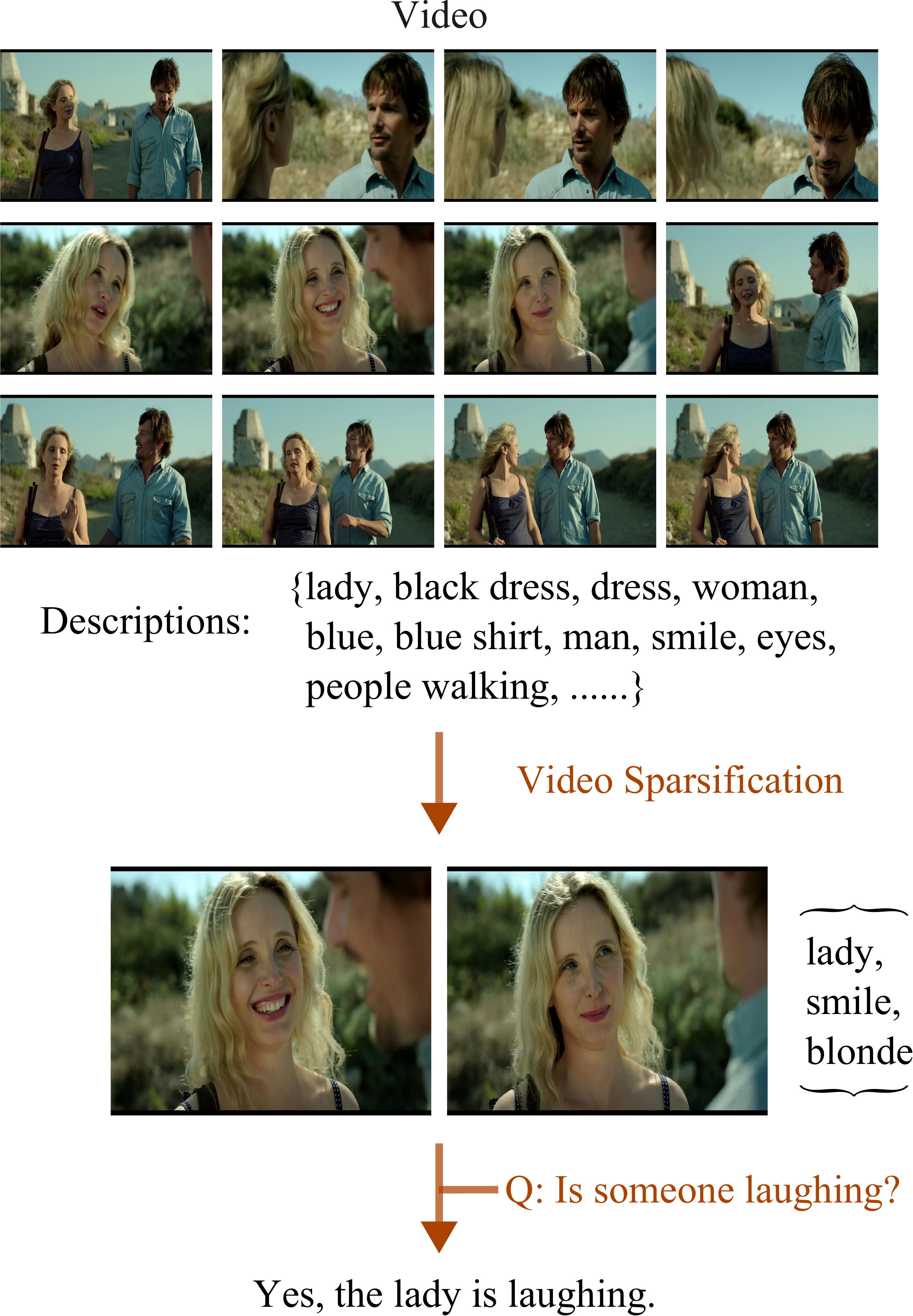}
    \caption{We study the Sparsified VideoQA problem, where we learn to sparsify the original long video into very few inputs for QA. We design a video sparsification process to deal with video of multiple modalities (frames, word and phrase descriptions, etc). }
    \label{fig:teaser}
\end{figure}

Watching long videos is time-consuming and easily loses user attention. How to efficiently present videos to users is an important and practical problem in various video applications. For example, for home surveillance videos which are usually recorded continuously throughout the day, it is hard for users to capture a moment of package delivery from an hour-long video. More generally speaking, for videos that are not carefully edited (e.g., Youtube videos), they often contain purposeless parts and need pre-processing of content so that users can quickly get meaningful information.

Videos often come from different modalities. Commonly, they are composed of image frame sequences. With the advances of recording devices and editing tools, videos often contain speech (e.g., Youtube videos recorded from user phones) and subtitles (e.g., in movies and TV-shows). It has been shown that leveraging different modalities benefits various video tasks~\cite{Yang_2021_ICCV, li2020hero}. However, it is worthwhile noticing that the various modalities in video could be quite noisy and redundant --- meaningless utterances, repeating frames, etc.\ --- causing computational inefficiency and distracting model learning. Furthermore, the problem of modality imbalance~\cite{goyal2017making} has been studied, where the unbalanced information across modalities could result in significant bias towards one modality. For example, prior works~\cite{lei2019tvqaplus} have shown that in TV-related videos, the major contribution for the various video-language tasks comes from subtitles while the video frames play a negligible role. 

In this work, we characterize the VideoQA problem from the perspective of input sparsity. As illustrated in Figure~\ref{fig:teaser}, we aim to answer the question: ``How much visual/textual signals are sufficient for a task?'' For VideoQA specifically, different questions require different amount of video information to give the answer. For example, if the question asks about people, then theoretically the system only needs to look at the moments where people are present. In the literature, there is evidence showing that video action classification can be accomplished with single frame~\cite{wu2019adaframe, huang2018what}. Recently there have also been works that imply sparse uniform sampling of the video is sufficient for video and language tasks~\cite{lei2021less}, and an analysis tool which shows that video and language tasks could be achieved by picking one optimal frame~\cite{buch2022revisiting}. In this work, we instead move beyond single frame input, and try to characterize the role of videos by learning to select an optimal set of video inputs. We propose a generic framework which learns to drop video inputs while training for the video-and-language task. This framework can be applied to different kind of video modalities, and in our experiments we provide analysis on visual-only (i.e., video frames), text-only (i.e., video subtitles or key words), and visual-textual inputs. 

From our experiments, we demonstrate that with very sparse inputs, the task can still be accomplished pretty well. Specifically, we are able to achieve $5.2\% {-} 5.8\%$ loss of accuracy with only $10 \%$ length of the video, which corresponds to only $2{-}4$ selected frames. Meanwhile, we also observe complimentary behaviour between modalities even with sparsified multi-modal inputs. Our work suggests the potential of improving data efficiency under either single or multi-modal settings for video-and-language tasks.

\section{Related Works}
\noindent\textbf{Video Question Answering}
VideoQA is a video understanding task about predicting answers given a video and a related question~\cite{Zhu2017UncoveringTT,jang2017tgif,xu2017video,lei2018tvqa,tapaswi2016movieqa}. Recent works usually use multi-modal transformer models~\cite{li2020hero,lei2021less} and feed in a combined sequence of frame and question word tokens, and the model processes the multi-modal inputs with attention mechanisms. To ensure full utilization from a video, prior works usually sample dense frames and feed them altogether into the model, trying to maintain as much information from the video as possible, and then perform analysis based on the assumption of minimal information loss from the video at the feature level. We instead want to characterize the video behaviour of VideoQA task from the perspective of limited information.

\noindent\textbf{Vision vs.\ Language in Multi-modal Learning}
It has been pointed out by several works \cite{lei2021less, huang2023video} that there exists information imbalance between different modalities. Especially, \cite{goyal2017making} shows the language bias in image-language tasks and provides a solution to overcome it. Recently, \cite{huang2023video} shows that video-language tasks can be accomplished pretty well with as little as 10-bit of information, again implying the strong bias towards language in video-language tasks. Recently, \cite{buch2022revisiting} provides a detailed analysis showing video-language tasks weakly rely on the temporal information from video and that video-language tasks can be comparably accomplished by a strong image-language baseline. Our work is also built upon these interesting observations and aim to characterize the behaviour of multi-channel video inputs in different downstream tasks.

\noindent\textbf{Efficient Video Understanding}
Our work is related to some recent works on improving video data efficiency in video understanding applications. \cite{lei2021less} shows that sparsely sampled video frames can be trained end-to-end to perform video-language tasks well. \cite{wu2019adaframe} study how to reduce video frames for fast video recognition tasks. There are some evidence~\cite{buch2022revisiting, huang2023video} showing video understanding tasks can be accomplished even with reduced inputs. Prior works~\cite{buch2022revisiting, wu2019adaframe} often focus on extracting single key frames to represent video. \done{The previous two sentences need citations} More recently, \cite{wang2022efficient} proposes efficient vision transformer for video action recognition by dropping spatial-temporal tokens.
We offer a way to analyze video-language tasks, which typically rely on transformer models, with inputs at user-controlled sparsity levels. Given the multi-channel nature of videos, we can similarly study textual-based sparse representation of videos as well.

\section{Approach}
In this section, we introduce our token sparsification approach. We first provide a brief preliminary on multi-modal transformers. Then we explain how the learnable token sparsification works. Finally, we explain how we can extend it to multi-modal setting. 
\done{text summarizing this section}
%\gunnar{Is it possible to do some sort of a model figure to explain the architecture etc?}
\subsection{Preliminaries --- Multi-modal Transformers}
\done{text summarizing this section}
%\noindent\textbf{Multi-modal Transformers}
To solve for video tasks that involve multi-modal inputs (e.g., VideoQA), current literature usually converts both visual and textual signals into sequence of embeddings, projects them into a common embedding space and applies a multi-modal transformer which takes the sequence concatenation to generate the final output. i.e., 
\begin{equation}\label{eq:task_perform}
    y = \cM([v_1, \dots, v_n; w_1, \dots, w_m]) \,,
\end{equation}
where $\cM$ is the multi-modal transformer, $y$ is the task output (e.g., predicted answer in text), $\{v_i\}_{i=1}^n$ is the sequence of video frame/segment features, and $\{w_j\}_{j=1}^m$ is the sequence of word embeddings for the text inputs. 

To get $\{v_i\}$, the common practice is to span the entire video, and use an off-the-shelf feature extractor to get frame or segment level features (depending on whether the feature extractor is image-based or segment-based). In this way, all the video information is included for the task. However, videos contain a lot of redundant information which is not necessarily useful for the task. For example, if the task is asking about ``What the person is wearing'', intuitively it only needs one video frame that contains the target person. From this perspective, we aim to design a framework that learns to select only the key information for the task.

\begin{figure}[t]
    \centering
    \includegraphics[width=\linewidth]{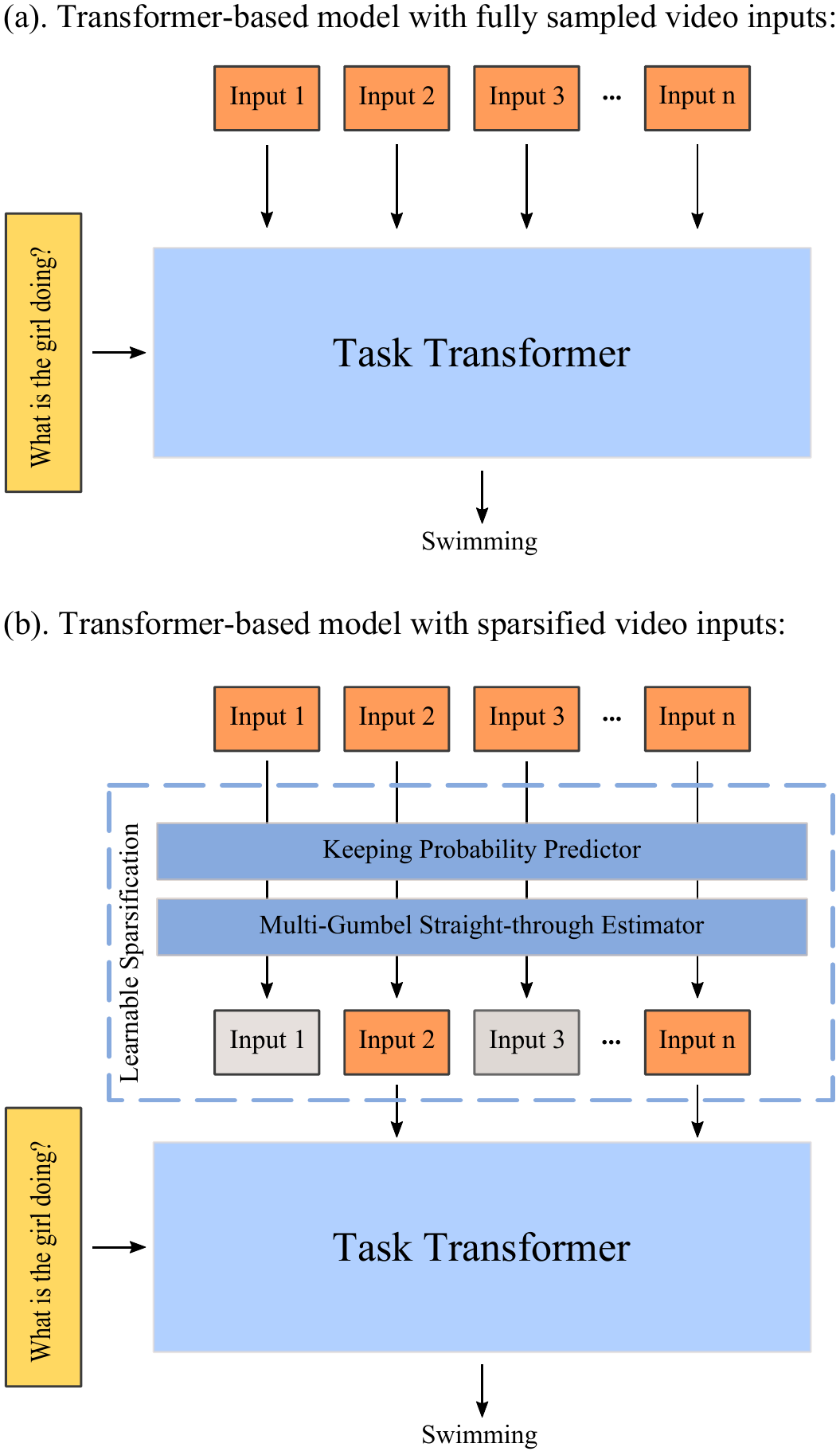}
    \caption{The pipeline of our method. We insert a learnable selection module between the inputs and the task transformer to sparsify the inputs. Our method can be generally applied to single or multi- modal inputs from videos. The entire network is trained end-to-end using the task objective as well as the input sparsity constraints.
    \done{iterate on teaser figure. we probably want to show the multimodality? Also, we want to show that this is using a video input?}}
    \label{fig:concept}
\end{figure}

\subsection{Token Sparsification}\label{sec:sparsification}
Observing the architecture of transformers which treat visual and/or textual embeddings as sequence tokens, we propose to learn to sparsify the input tokens during training. Starting with single modality inputs $\{v_i \}_{i=1}^n$, we first generate the keeping probability $s_i$ for each token $v_i$ to estimate how likely $v_i$ should be kept. 
To ensure $s_i$ is a valid probability value within the range of $[0,1]$, we place a predictor $f(\cdot): \cR^d \rightarrow \cR^2 $ to map each token $v_i$ to 2-dim, and apply a Softmax normalization to re-scale the 2-dim vector into $[0,1]$, and take the first entry as the keeping probability $s_i$. Specifically,  
\begin{equation*}
s_i = < \begin{pmatrix} 1\\0 \end{pmatrix},  \text{Softmax}(f(v_i))>, % \in \cR^2 ,
\end{equation*} \done{why is the output two-dimensional if $s_0$ is just the scalar probability?}
%where $s = [s_0, s_1]$. $s_0$ is the first entry of vector $s$ which represents the probability that token $v$ should be kept. \done{token $v$ or token $v_0$?} 
where $<\cdot>$ is the inner product function. At inference time, we can treat sparsification as a ranking procedure, whereas we rank tokens according to their keeping probabilities and select top-K tokens. 

During training, to facilitate the learnability of the network, we treat sparsification as a top-K sampling procedure. That is, we draw K samples from $\{ v_i \}$ from their keeping probabilities as the sparsified inputs to the rest of the model. By default, the sampling process is not differentiable. In order to overcome this non-differentiability, we refer to and extend the Gumbel-Softmax~\cite{jang2016categorical} trick. 

\noindent\textbf{Gumbel-Softmax Straight-through Estimator}~\cite{jang2016categorical} offers a differentiable way for \textit{single} discrete sampling. It first re-parameterizes the sampling distribution by adding Gumbel noise followed by a Softmax normalization, i.e., 
\begin{equation*}
s_i^g = \frac{exp((\log s_i + g_i)/\tau)}{\sum_{j=1}^n exp((\log s_j + g_j)/\tau)} \, ,
\end{equation*}
Then it follows the design of straight-through estimator~\cite{toderici2015variable} to  get the sample index from $k = argmax_i \{s_i^g\}_{i=1}^n$ and select the corresponding $v_k$ in the forward pass while zeroing-out the rest. In the backward propagation, %it follows the gradient of $\frac{\partial \sum_j s_j^gv_j}{\partial s_i}$. 
the gradients of $s_i^g$ are kept for use.
We encourage referring to~\cite{jang2016categorical} for more details and proofs.  

\iffalse
\noindent\textbf{Loss-controlled Gumbel-TopK STE (Loss-Gumbel-TopK STE)}.
The standard Gumbel-Softmax STE trick only supports single sampling, as  it performs along the sequence and only selects one most likely token. In our context, we'd like to extend it to multi-sampling of top-K most likely tokens. A Gumbel-TopK trick is introduced by~\cite{kool2019stochastic} that performs sampling without replacement. We'd like to further reduce the complexity of multiple sampling by doing sampling once while imposing an auxiliary loss to control the sparsity. Our approach also doesn't require deciding $K$ at training time, which adds more flexibility. 
In detail, we use Gumbel-Softmax trick on top of each token's keeping distribution, i.e, we can get
\begin{equation*}
    s_i^g = \frac{exp((\log s_i + g_{i_0})/\tau)}{exp((\log s_i + g_{i_0})/\tau) + exp((\log (1-s_i) + g_{i_1})/\tau)} 
\end{equation*}
where $g_{i_0}, g_{i_1} \sim Gumbel(0,1)$ are Gumbel noises. We keep all the tokens whose perturbed keeping probability $s_i^g > 0.5$. 
\fi 

\noindent\textbf{Multi-Gumbel Straight-through Estimator}. The standard Gumbel-Softmax STE trick only supports single sampling, as  it performs along the sequence and only selects one most likely token. In our context, would like to extend it to multi-sampling of  most likely tokens. ~\cite{kool2019stochastic} introduces a trick to perform sampling without replacement. In our implementation and experiment, we instead modified ~\cite{kool2019stochastic} to two variants:
\begin{enumerate}
    \item \textbf{Gumbel-TopK Selection}: select the top-K tokens with higher $s^g_i$ values. And then use straight-through estimator after selection, i.e., we only keep the selected tokens and zero-out the rest, but in the backward pass, we still use the gradients of all $s^g_i$. 
    \item \textbf{Ratio-controlled Gumbel}: instead of hard selection of K samples, we allow sampling with arbitrary number of tokens while keeping a sparsity ratio constraint in the loss during training. Specifically, we add Gumbel disturbance to $s_i$:
    \begin{equation*}
    s_i^g = \frac{exp((\log s_i + g_{i_0})/\tau)}{exp((\log s_i + g_{i_0})/\tau) + exp((\log (1-s_i) + g_{i_1})/\tau)} 
    \end{equation*}
    where $g_{i_0}, g_{i_1} \sim Gumbel(0,1)$ are Gumbel noises. We then keep all the tokens whose perturbed keeping probability $s_i^g > 0.5$. For a human-selected target keeping ratio $p$, we add a loss constraint on the overall keeping ratio over the batch:
    \begin{equation*}
        \cL_{select} = \frac{1}{B}\sum_{b=1}^B(p - \frac{1}{n}\sum_{i=1}^n\mathds{1}[v^b_i \text{ is kept}])^2 ,
    \end{equation*}
    where $B$ is the batch size, and $v^b_i$ is denoted as the tokens within a batch.
\end{enumerate}
The overall training loss generically is the weighted balance between the task loss and the selection loss:
\begin{equation}\label{eq:loss}
\cL = \cL_{task} + \lambda \cL_{select},
\end{equation}
where $\lambda$ is the balancing weight between two loss components, and $\cL_{task}$ is the task loss. Note that $\cL_{select}$ is not necessary required for the first variant and we can simply set $\lambda=0$ in that case. The method is generic and task-agnostic. For VideoQA, we refer to~\cite{li2020hero} for $\cL_{task}$, which essentially is a cross entropy loss between the predicted answer and the ground-truth answer word.   
%\gunnar{define $\lambda$}
During inference, we directly rank $\{s_0 \}$ and select the inputs associated with the top-K scores. In this way,  we are able to get the sparsified video inputs for the target task.

\subsection{Sparsified Positional Encoding}
Positional encoding is a typical mechanism designed for transformers to encode the order of tokens within a sequence. Typical designs for positional encoding are either fixed, such as sinusoidal function with selected frequencies~\cite{vaswani2017attention}, or learnable positional embeddings like~\cite{dosovitskiy2020image}. In our context, we make use of learnable positional embeddings following~\cite{li2020hero, Yang_2021_ICCV}. For a full-length sequence, each token $v_i$ is added to a unique learnable embedding vector $p_i$ that sticks to this position $i$. After sparsification, tokens are dropped and kept by probability, so the remaining tokens could be at various positions. 

In our preliminary experiments, we found that inappropriate positional embedding could harm training convergence, especially when we just leave the positional embeddings as they are in the full sequence scenarios. As a result, in our implementation, we arrange the sparsified tokens $\{v_{k_i}\}_i^K$ into a new sequence $\{v'_{i}\}_i^K$ (where $v'_i = v_{k_i}$), and assign $p_i$ to $v'_i$ accordingly. We observe a good convergence behaviour and performance from this simple rearrangement.

%\subsection{Question-conditioned Pruning Ratio Control}

\subsection{Multi-modal Token Sparsification}\label{sec:multimodal}
We generalize the standard single modality token sparsification to multi-modal scenarios where videos come with both visual and textual signals. Specifically, our method allows the inputs being any kind of tokenized inputs. For videos that come with multi-channel inputs, e.g., subtitles, we can directly concatenate time-stamped subtitles along with the frames as multi-modal inputs. Then we can perform similar token sparsification on both of them. Specifically, we denote the multi-modal inputs as $\{v_i, w_i\}_{i=1}^n$ (note that we can assume same number of $v_i$'s and $w_i$'s by interpolating or padding, etc). We first get the unified representation from multi-modal inputs by
\begin{equation}
    u_1, \dots, u_n = \cC_m([v_1, \dots v_n; w_1, \dots, w_n])
\end{equation}
where $\cC_m$ is a context model (e.g., cross-modal transformer) that exploits both visual $v_i$ and subtitle $w_i$ information. Then we apply the uni-modal sparsification technique introduced in Sec.~\ref{sec:sparsification} on top of $\{u_i \}$ sequence to compute the keeping scores, and sparsify the original multi-modal input sequence $\{ v_i, w_i \}$ according to the score, to get the sparsified multi-modal sequence $\{v_{k_i}, w_{k_i} \}_{i=1}^K$. Then this sparsified sequence is fed to the task performer $\cM$ to compute the outputs (eq.~\ref{eq:task_perform}).

We also consider another approach where we ease the restriction on time-matched input pairs and operate on each modality regardless of their timestamps. In this setting, we apply different context model on each modality sequence, and then use separate selection loss to constrain the sparsity. We will provide more details on this setting in the Experiment details, where we demonstrate our idea with key-word selection that summarizes the videos.  
%-------------------------------------------------------------------------

\begin{figure*}[h]
    \centering
    \includegraphics[width=0.9\linewidth]{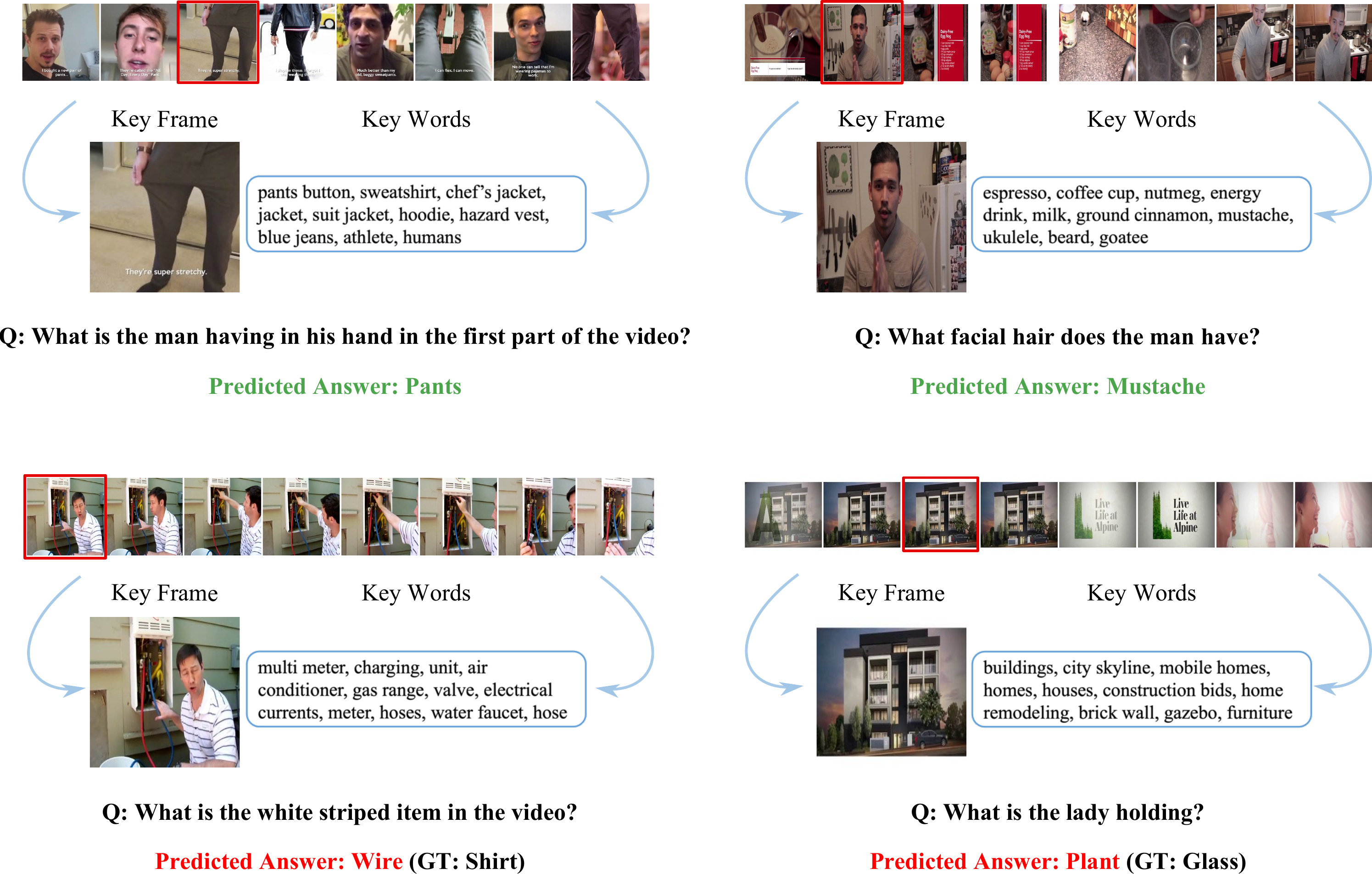}
    \caption{Qualitative examples on iVQA videos. Single frame and ten-word summary is generated from the original video for Video Question Answering task. First two examples demonstrate successful cases where both visual and textual signals signals are able to capture the question-relevant information. The last two examples show some failure cases where visual and/or textual signals are distracted from the question.\done{Figure could be improved by clearer labels for what is going on and thicker lines, etc.}}
    \label{fig:vis_1}
\end{figure*}

\section{Experiments}
\done{section summary here}
In this section, we provide our experiment results. First we detail our implementation and VideoQA datasets; then we provide VideoQA results under different input sparsities, followed by multi-modal results. Finally, we offer some qualitative visualization to analyze our approach.

\begin{table*}[h]
\centering
\caption{Effect of the temperature $\tau$. Smaller $\tau$ leads to more exploitation while higher leads to more exploration. We observe that more explorative selection is beneficial for denser inputs.}
\label{tab:tau_ablation}
\begin{tabular}{cccc|ccc}
\toprule
{\ul }   & \multicolumn{3}{c}{VLEP}           & \multicolumn{3}{c}{VIOLIN}   \\
Input Percentage & $\tau = 0.01 $ & $\tau = 0.1 $ & $\tau = 0.5 $ & $\tau = 0.01 $ & $\tau = 0.1 $ & $\tau = 0.5 $ \\
\midrule
10\%     & \textbf{60.25} & 56.01          & 58.94          & 56.25    & \textbf{60.57} & 58.80          \\
30\%     & 60.95          & \textbf{63.52} & 59.13          & 61.72    & 57.64          & \textbf{62.34} \\
50\%     & 63.05          & 63.64          & \textbf{64.30} & 65.57    & 64.48          & \textbf{66.06} \\
70\%     & 63.73          & 65.14          & \textbf{65.32} & 65.94    & 66.52          & \textbf{67.06}
\\
\bottomrule
\end{tabular}
\end{table*}

\begin{table*}[h]
\centering
\caption{Effect of the balancing weight $\lambda$. $\lambda$ balances the selection loss and task loss as specificed in eq.~\ref{eq:loss}. We report results on VLEP dataset and observe that $\lambda=1.0$ yields the best balance. Higher $\lambda$ might lead to distraction of task, while lower $\lambda$ might lead to insufficient sparsification. We pick $\lambda=1.0$ based on the following ablation.}
\label{tab:lambda_ablation}
\begin{tabular}{ccccc}
\toprule
%{\ul }   & \multicolumn{4}{c}{VLEP}          \\
Input Percentage & $\lambda = 0.01 $ & $\lambda = 0.1 $ & $\lambda = 1.0 $ & $\lambda = 10.0 $ \\
\midrule
10\%     &  59.12    &   59.85   &  60.25       &   59.90    \\
70\%     &   65.23      &     65.32     & 65.32 &   65.11  \\
\bottomrule
\end{tabular}
\end{table*}

\begin{table*}[h]
\caption{Comparison of our two Gumbel variants. Overall the first variants perform slightly better. The second variant is superior at highly sparsified level ($10\%$) as it adds more flexibility in individual sparsity levels across different videos. }
\centering
\label{tab:variants}
\begin{tabular}{ccccc}
\toprule
%{\ul }   & \multicolumn{3}{c}{VLEP}          \\
Input Percentage & $10 \% $ & $ 30 \% $ & $ 50 \% $ & $70 \% $  \\
\midrule
Gumbel-TopK Selection     &  60.25   &  63.52    &  64.30  &   65.32     \\
Ratio-controlled Gumbel     &  61.43   &  63.42    &  63.49  &  65.01 \\
\bottomrule
\end{tabular}
\end{table*}

\begin{figure*}[h]
    \centering
    \includegraphics[width=\linewidth]{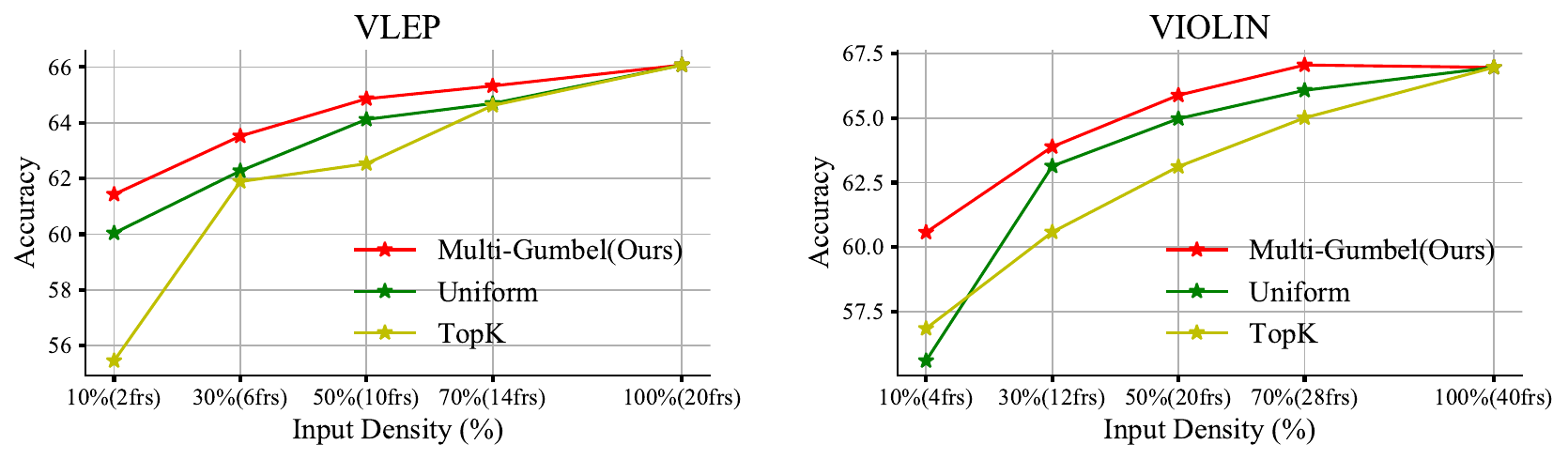}
    \caption{Sparsified VideoQA results on VLEP and VIOLIN datasets. Accuracy at the $100\%$ level refers to the original full input baseline result. We can conclude that learnable sparsification is better than fixed sampling (Uniform), and that stochasitic sampling is better than deterministic selection (TopK). Our Multi-Gumbel estimator achieves the best result overall.}
    \label{fig:qa_curve}
\end{figure*}

\begin{table*}[h]
\centering
\caption{VideoQA results on iVQA. We apply our approach on the state-of-the-art method~\cite{Yang_2021_ICCV}. We consider multi-modal sparsification where we sparsify both visual (i.e., frames) and textual (i.e., words) inputs. Compared to single-modality, multi-modal performance is stronger at different sparsification levels. With additional extracted words, we also outperform the state-of-the-art result on iVQA (last column).}
\label{tab:ivqa}
%\resizebox{0.9\textwidth}{!}{
\begin{tabular}{cc|ccccc}
\toprule
\multicolumn{1}{l}{}                                                       &           & \multicolumn{5}{c}{Visual (Snippets)}                             \\
\multicolumn{1}{l}{}                                                       &           & 0             & 1 snippet & 2 snippets & 5 snippets & 20 snippets \\
\midrule
\multirow{5}{*}{\begin{tabular}[c]{@{}c@{}}Textual\\ (Words)\end{tabular}} & 0         & 14.6 (Q-only) & 28.65     & 30.24      & 31.26      & 35.43~\cite{Yang_2021_ICCV}       \\
& 5 words   & 17.5          & 28.68     & 30.31      & 31.70      &    35.43         \\
& 10 words  & 18.22         & 29.87     & 31.43      &  31.88      &    36.01         \\
& 25 words  & 20.14         & 30.16          &   31.59         & 32.03      &  36.09           \\
& 100 words & 26.75         & 31.47          &   32.11         &  33.21          & \textbf{36.42} \\
\bottomrule
\end{tabular}
%}  % resizebox
\end{table*}

\subsection{Implementation Details}
To verify our idea, we experimented on two state-of-the-art video-and-language models VQA-T~\cite{Yang_2021_ICCV} and HERO~\cite{li2020hero}. HERO considers multi-channel videos where videos come with subtitles as additional channel of inputs. HERO follows a hierarchical transformer architecture to first exploit the information within video modalities and contexts, and then has another task head to operate the task. VQA-T simply consists of two Distill-BERT models to deal with video+question inputs and answer candidates, and computes the answer based on embedding similarity.
For extracting the video features, we follow~\cite{Yang_2021_ICCV} to use the S3D model pre-trained on Howto100M dataset. For extracting the key word candidates, we use the model offered by~\cite{Yang_2021_ICCV} and the vocabulary from the training split of the dataset to extract the words/phrases based on feature similarity.

\subsection{Datasets and Metrics}
We evaluate our idea on public VideoQA benchmarks including VLEP~\cite{lei2020vlep}, VIOLIN~\cite{liu2020violin} and iVQA~\cite{Yang_2021_ICCV}. For VLEP and VIOLIN, we follow~\cite{li2020hero} to build our method on top of HERO. VLEP and VIOLIN provide both raw videos and subtitles as inputs. Our selection is then based on the multi-modal inputs. For iVQA, we follow~\cite{Yang_2021_ICCV} to build our method on top of VQA-T. 
We report VideoQA accuracies across different input sparsity level: $10\%$, $30\%$, $50\%$, $70\%$ and full ($100\%$) inputs.

\subsection{VideoQA Experiments}
\done{Introduction for this experiment, what is it meant to show?}
%First, we study the trade-off between task performance and the input sparsity, in order to verify our learned sparsification as well as observe how many inputs the task acquires. 
We present our single modality sparsified VideoQA results here. First, we study the design choices of our two multi-gumbel estimator variants, followed by the comparison between our approach and other token sparsification baselines.

\noindent\textbf{Effect of temperature $\tau$}. In our experiments, we found that varying $\tau$ could result in very different performance. We elaborate the result in Table~\ref{tab:tau_ablation} with our Gumbel-TopK selection variant, where we choose $\tau = (0.01, 0.1, 0.5)$ for each sparsity level, and fix $\lambda=1.0$. A smaller $\tau$ means the model focuses more on exploitation, while a larger $\tau$ makes the model focus more on exploration. We can observe that on both datasets, the model that is more explorative with denser inputs gives a better results; but on sparser inputs, the model tends to stick to exploitation.

\noindent\textbf{Effect of loss balancing weight $\lambda$}. We also study how the balancing weight $\lambda$ affects the performance with our ratio-controlled Gumbel estimator. In Table~\ref{tab:lambda_ablation}, we choose $\lambda=0.01, 0.1, 1.0$ and $10.0$, and then report the results at highly sparsified ($10\%$) and lowerly sparsified ($70\%$) levels on VLEP dataset. We fix $\tau=0.01$ for $10\%$ level and $\tau=0.5$ for $70\%$ level. $\lambda$ has slightly larger impact on highly sparsified setting.
We observe that $\lambda=1.0$ yields the best balance and hence choose it for all the other performance.

\noindent\textbf{Comparison of two multi-gumbel variants}. We compare the two variants of Gumbel estimator for token sparsification. In Table~\ref{tab:variants}, we compare these two variants at different sparsity levels on VLEP. Our Gumbel-TopK selection variant is better than ratio-controlled Gumbel overall. Ratio-controlled Gumbel is superior at highly sparsified level ($10\%$) as it adds more flexibility in individual sparsity levels across different videos. 

\noindent\textbf{Comparison with other sparsification approaches}. To our knowledge, no prior work has studied the same topic on VideoQA before, so there is no direct comparison. To validate our approach of Multi-Gumbel Estimator, we define the baselines on our own:
\begin{enumerate}
    \item Uniform(Fixed): Fixed uniform sampling of inputs w.r.t. different sparsity levels. 
    \item TopK: During training, directly select inputs with higher keeping probability $s_i$ after softmax step, without noise perturbing.
    \item Multi-Gumbel(Ours): Our approach which stochastically sparsifies tokens with Gumbel perturbing, we plot the better result from the two variants we introduced. 
    %\item Gumbel-TopK~\cite{kool2019stochastic}: During training, draw top-K samples without replacement (i.e., repeat Gumbel STE for K times and gather the K different tokens); note that this requires prior knowledge of K. We select $K=10, 20$ respectively for VLEP and VIOLIN datasets.   
\end{enumerate}
We show the accuracy vs.\ density curve in Figure~\ref{fig:qa_curve}. We can see that our Multi-Gumbel approach module is able to achieve the best performance across different sparsity levels. Compared to learnable selection, fixed uniform sampling is weaker as it does not contain any form of task adaptive selection. A direct TopK selection training performs weaker than training with stochastic sampling, as we observe that the deterministic selection tends to a local optimal choice, while our stochastic Multi-Gumbel approach gives more flexibility of by adding noises while learning. One noticeable observation is that, at $10\%$ level, which corresponds to very few frames (2 frames for VLEP and 4 frames for VIOLIN), the performance is still quite good. It implies the potential of accomplishing the task with very few inputs.

\begin{figure*}[h]
    \centering
    \includegraphics[width=\linewidth]{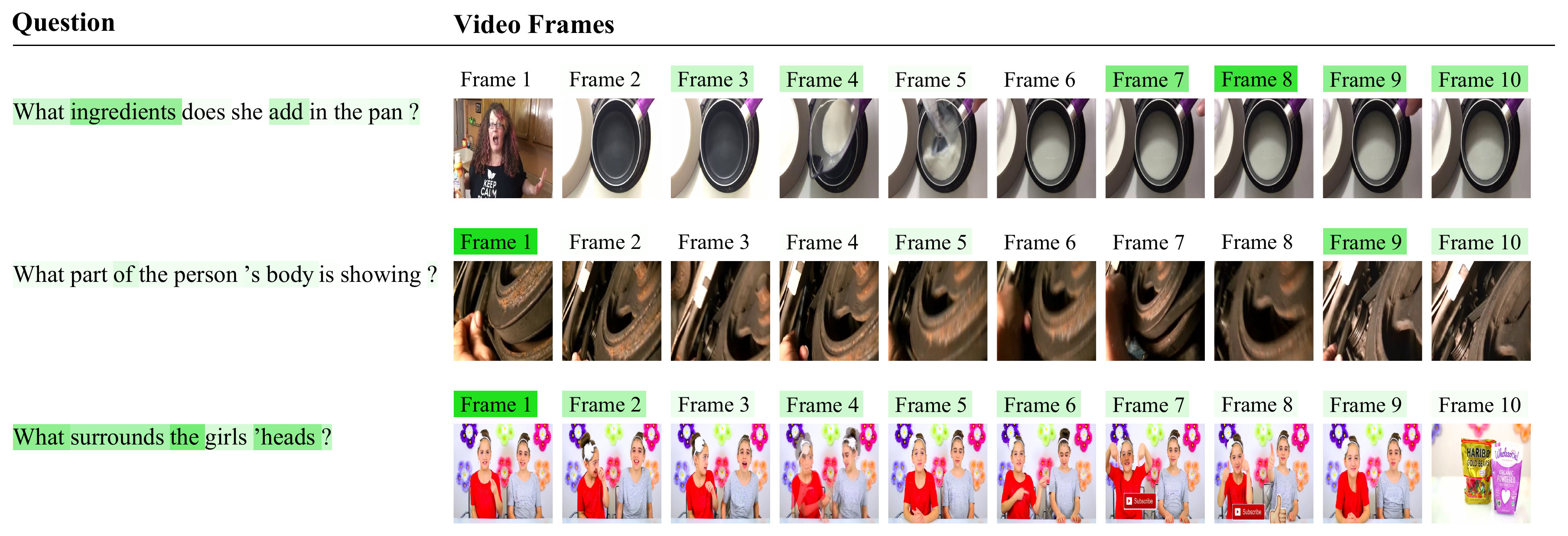}
    \caption{Frame importance visualization. Darker color means the corresponding word/frame is of more importance to predict the answer. We can see that the model is able to discard some repetitive frames or frames that are not relevant. \done{Add labels to the image? I.e. Question, video, etc?}}
    \label{fig:vis_2}
\end{figure*}

\subsection{Multi-modal Sparsification Results on iVQA}
In the multi-modal experiments, we would like to study the relation between visual and textual modalities under a controlled input setting. In order to do that, we extend our learnable selection module to the multi-modal setting following Section~\ref{sec:multimodal} to generate key frames and key words from the original video inputs. 
We first get a pool of candidate inputs from the raw video. The candidate frames are directly sampled from the videos, while the candidate key words are extracted using CLIP-based model, which finds the closest words or phrases using nearest embedding matching. We use all the phrases and words from the iVQA training set as the vocabulary dictionary to choose words from. To better demonstrate the results, we use the format of few-word or few-frame inputs. For visual frame inputs, we process with the same method as before. For textual inputs, we treat 5-word as one unit. 5-word/sec is the average reading speed for adults, which consumes similar attention from watching a frame. So 5-word and one single frame could be thought of as equivalent in consuming user attention. We combine the units into a sequence, then apply the same selection method for word selection. For multi-modal setting, we concatenate the frames and word units as a multi-modal sequence and select from both. We fix $\tau = 0.1$ for training the models.

Our results are shown in Table~\ref{tab:ivqa}. For single-modality inputs, we similarly observe an increasing performance trend with increasing number of inputs. Even with very few inputs, the VideoQA performance is very close to the upper bound from dense inputs. We can also observe a boost of performance from increasing density of inputs on both modalities at sparsified levels, which validates the effectiveness of our sparsification techniques. In general, the visual inputs perform stronger than textual inputs, which is mainly due to the fact that visual signals are much more informative. On the other hand, we can still observe an increase of performance from adding even very few multi-modal inputs. For example, adding only 5-word to the visual snippet could still get some performance gains. This implies the complimentary manner from different modalities from the perspective from strictly controlled inputs. Noticeably, as an intermediate output from our learnable selection, we can get a few-frame and few-word summarization of the original video, which is human-interpretable. We provide more examples and analysis in the following section to demonstrate this advantage.

\subsection{Qualitative Analysis}
\done{figure 4 is mentioned before figure 3}
Here we provide visualizations on the selected frame and/or key words from iVQA dataset. For illustration purposes, we present the result for single frame selection and 10-word extraction in Figure~\ref{fig:vis_1}, along with their associated questions and the predicted answer. Answer in green color means the system correctly predicts the answer, while red color means the system predicts the wrong answer and the ground truth is in parenthesis. In the successful cases, the sparsified output is able to capture an appropriate figure for the topic, and the texts also contain words related to the answer, which leads to the correct answer. In the first failure case, even though the selected frame contains information related to the answer ``shirt'', the textual component is a distraction, and the system generates an answer more related to the key words which are closely describing pipes. In the second failure case, the generated key frame and words are both irrelevant to the question. This is probably because the question itself is asking something minor (since most of the video contents are about the architectures and surroundings) while the model is trained to get information that is of major interest for the overall dataset and task. 

Additionally, we analyze the token importance using the tool provided by~\cite{Chefer_2021_CVPR} which calculates the importance score of each input token w.r.t. to the task prediction. In Figure~\ref{fig:vis_2}, we provide some visualization examples where question words and video frame inputs are highlighted according to their importance scores. For illustration purposes, we only sample 10 frames in each sample. Words or frames that are of darker green color means they contribute more to the prediction. From the given examples, we can see that not every video frame is of significant importance. The model is able to discard frames that do not contain any useful information for the question (e.g., in the second example, only the frames showing the fingers are contributing). On the other hand, in the example where the scene is relatively stable, we can also observe that the model focuses mostly on one of these similar frames (as in the third example), while the rest seems to be diverging. These observations show the potential of dropping unnecessary video inputs to improve the efficiency, which validates our motivation.

%-------------------------------------------------------------------------
%\subsection{Video Summarization Experiment}
%TODO: put results on Wikihowto 

%-------------------------------------------------------------------------
%\subsection{Video Summarization Experiment}
%TODO: put results on Wikihowto 

\section{Conclusion}
In this work, we characterize video question answering from the perspective of sparsified inputs. We propose to use a learnable selection module to adaptively select the best and representative input. This allows us to get multi-length input on different types of modalities. In our experiments, we analyze the current Video Question Answering benchmarks, where we observe a fair performance with a small input budget. We also observe a complimentary performance under a multi-modal setting. We believe our work meaningfully shows the potential of improving data efficiency under various video representation types.

%%%%%%%%% REFERENCES
{\small
\bibliographystyle{ieee_fullname}
\bibliography{egbib}
}

\end{document}